\documentclass[a4paper]{article}

\usepackage{algorithm}
\usepackage{algorithmic}
\usepackage{tabularx}
\usepackage{url}
\usepackage{graphicx}
\usepackage{amsmath}

\title{ svcR: An R Package for Support Vector Clustering improved with Geometric Hashing applied to Lexical Pattern Discovery }
\author {Nicolas Turenne  \\ INRA,  Paris}
\date{}

\begin{document}

\maketitle 

\begin{center}
INRA, Unit\'e Math\'ematique Informatique et G\'enome, UR1077 \\
F-78350 Jouy-en-Josas, France \\
E-mail: nicolas.turenne@jouy.inra.fr \\
URL: http://genome.jouy.inra.fr/~turenne/ \\
\end{center}

\begin{abstract}
 
  We present a new R package which takes a numerical matrix format as data input, and computes clusters using a support vector clustering method (SVC). We have implemented an original 2D-grid labeling approach to speed up cluster extraction. In this sense, SVC can be seen as an efficient cluster extraction if clusters are separable in a 2-D map. Secondly we showed that this SVC approach using a Jaccard-Radial base kernel can help to classify well enough a set of terms into ontological classes and help to define regular expression rules for information extraction in documents; our case study concerns a set of terms and documents about developmental and molecular biology.
\end{abstract}


\section[Introduction]{Introduction}\label{sec:svc}
Mining text archives is a great challenge since lots of documents are available and their amount grows in the same way as the capacity of computer storage. Making rules of a domain for knowledge extraction involves efficient features with low semantic ambiguity. It is not an easy task and we try to answer this question by representing vectors of linguistic expressions (i.e. terms) by features and using a scalable density-based distance to cluster the terms.   \\ 
The first idea for our problem concerns the choice of a density-based method and the improvement of its scalability. Clustering can be a useful knowledge-poor technique to induce organization into scattered data \cite{Jain:1988}. Non-parametric methods such as support vector machines can be interesting to analyze noisy data by density processing. \cite{BenHur:2001, Schölkopf:2001, Horn:2001} proposed an unsupervised support vector algorithm to enclose data clusters by contours and based it on a radial kernel. Diverse applications have been tested for novelty detection \cite{Eskin:2002, Lazarevic:2003}, rule extraction \cite{Zhang:2005}, désoxyribo-nucleic acid (DNA) and chemical compounds \cite{Bilen:2005, Eveillard:2002} or image processing \cite{Campedel:2005}. The method of point assignation to contours and related clusters is based on adjacent points between each pair and is time-consuming. Some studies \cite{Park:2004, Puma-Villanueva:2005} have been proposed to speed-up the method. In particular \cite{Yang:2002} and \cite{Lee:2005} proposed an improved method to label clusters, i.e. to assign point to clusters by graph analysis.   \\ 
We present a new robust method based on the computation of a hash function through surrounding points working with a grid which we map to data using a k-nearest neighbor method. We developed this clustering method under the R platform \cite{R-project:2010}, as a package called svcR, and we compared our approach to other ones, especially graph-based, on the Iris dataset (svcR is available from the Comprehensive R Archive Network at \url{http://CRAN.R-project.org/package=svcR}"). \cite{Kees:2003} have also developed a support vector method for clustering but using a divisive way iteratively searching a classical hyperplan separator based on classical support vector machine. The first step tries to separate the data set and a set artificially build in the same space of attribute values and the same size than the data set which is theoretically not justified; it seems that if not many classes are present (2 or 3) and not many attributes describe data, the algorithm seems to find groups, in other cases  it tends to find a number of final clusters equal to the number of iterations.   \\ 
The second idea presented in this paper concerns our original usage of support vector clustering (SVC) clustering methodology cited above for solving a certain form of ambiguity in natural language. Information retrieval \cite{Salton:1983, Daille:2003} and information extraction \cite{Kushmerick:2000, Soderland:1999} are key methodologies to retrieve information from text archives. But simple keywords may have several senses and assignment of term to conceptual classes should be important \cite{Gale:1992, Hearst:1992, Riloff:1993, Grefenstette:1994, Fellbaum:1998}. Clustering may be used to reduce the number of variables to take into account in rules for information mining in documents. We base our assumption on two works. Firstly that collocation analysis is useful to understand morphological structure and its link to a conceptual space \cite{Harris:1968, Smadja:1990}. Secondly that clustering can bring a good approach to build semantic classes with the help of a distance of similarity \cite{Pereira:1993, Nazarenko:1997}. This methodology about clustering linguistic terms can help to get common features to build rules for information mining in document archives. Classification of a set of terms requires to represent data as morphological information vectors (terms themselves or parts of terms and how?) and to determine which kernel has to be used to achieve SVC. We try to use whole terms, morphological primitives and bigrams as morphological information. And we try to use the Levenshtein distance and the Jaccard similarity index, Radial basis function, and combination Levenshtein-Radial or Jaccard-Radial kernels to study the clustering effect.   \\ 
In Section~\ref{sec:svc}, we introduce the methodology of support vector clustering. Section~\ref{sec:ghbl} presents the labeling approach and Section~\ref{sec:rep} gives studies of vector representation and different kernels for term clustering. Finally Section~\ref{sec:comp} shows evaluation of the technique.

\section[Support vector clustering]{Support vector clustering}\label{sec:svc}
In this section, we recall the clustering approach.
\subsection[Kernel trick]{Kernel trick}\label{subsec:fram}
We know a priori classes of items (red circles and yellow squares) and we search a linear frontier in a higher dimensional space. For that, data are transformed using a kernel function (dot product). Preprocess the data with:
 
\begin{eqnarray}
  \label{phi}
  \Phi:  X \rightarrow X \nonumber   \\  
       x \rightarrow x
\end{eqnarray}

$K$ is a dot product of the space (Hilbert space, $H$), and learn the mapping of  $\Phi(x)$ to $y$ (class).

\begin{eqnarray}
  \label{xi}
  \left\{ x_{i} \right\}^{N}_{i = 0}\ \mathrm{learning\ data\ x\ in\ X\ is\ a\ multivariate\ data\ on\ } X^{d}, \nonumber    \\ 
  \mathrm{ where\ d\ is\ the\ number\ of\ feature }
\end{eqnarray}

\begin{eqnarray}
  \label{K}
  \left\langle \Phi(x),\Phi(x')\right\rangle = K(x, x')\ \mathrm{can\ be\ computed\ in\ }X^{d} 
\end{eqnarray}

$\left( \left\| . \right\| \mathrm{is\ the\ norm\ associated\ to\ the\ K\ dot\ product} \right)$   \\ 
Usually $dim(X) << dim(H)$

\subsection[Optimization]{Optimization} \label{subsec:optim}
As an extreme view the distribution of data under the scope of unsupervised learning can be interpreted as density estimation. But in our case the approach estimates \textit{quantiles} of a distribution, not its density. In the case of SVC, we determine support vectors to delimit the distribution of points. The goal now points out to find the minimal sphere which surrounds data.
One can show: if $\Phi(x)$,...,$\Phi(x_{N})$ is separable from the origin in $H$, 
then the solution of margin minimization between two classes corresponds to the normal vector of the hyperplan separating the data from the origin with maximum margin.   \\  In our case we try to encapsulate data into a ball. The points inside the ball represent data to classify (first) and the origin represents the second class. Primal problem is written as follows. Let $a$ the (non-fixed) center of the ball, $R$ the radius of the ball and $C$ is a fixed penalty constant controlling the number of data near the ball. Let us minimize: 

\begin{eqnarray}
  \label{F}
  F\left( R, \left\{ r_{i} \right\}^{N}_{i = 1} \right)\ = R^{2} + C\sum_{i}r_{i}
\end{eqnarray}

Under the constraints:

\begin{eqnarray}
  \label{phia}
  \left\| \Phi(x_{i}) - a\right\|^{2} \leq R^{2} + r_{i}\ ,\ \mathrm{where}\ r_{i}\geq0\ \mathrm{for\ all\ }i = 1,..., N
\end{eqnarray}

$a$ is the center of the ball. The dual problem (for a convex problem) is the Lagrangian $L( R , a , \left\{ r_{i}\right\}^{N}_{i = 1}, \left\{\beta_{i} \right\}^{N}_{i = 1}, \left\{  \mu_{i} \right\}^{N}_{i = 1} )$. Where $\beta_{i}\geq0$ and $\mu_{i}\geq0$ are Lagrange multipliers, see \cite{BenHur:2001, Schölkopf:2001, Horn:2001} for details about computation of $\beta_{i}$. Multipliers are used to compute a (i.e. $\widehat{a}$) and $\Phi(x)$.

If $x$ is a support vector, the radius is:

\begin{eqnarray}
  \label{Rhat2}
  \widehat{R}^{2} = \left\|\Phi(x) - \widehat{a} \right\|^{2} 
\end{eqnarray} 

For any point $y$ the distance $Ry$ from the center is:

\begin{eqnarray}
  \label{R2}
  R^{2}_{y} = \left\|\Phi(y) - \widehat{a} \right\|^{2} 
\end{eqnarray} 

Hence it is possible to test if $y$ is inside the sphere or not by comparing $\widehat{R}$ and $R_{y}$.

\subsection[Contour deformation]{Contour deformation}
The value of parameter $\nu = \frac{1}{N.C}$ asymptotically represents a max bound of the BSV rate. Parameter $C$ takes values in $[\frac{1}{N}, 1]$, as $C$ is reduced, more and more points are labeled as outliers. If $q$ increases the Gaussian radius decreases and the number of SV increases. Subsequently if one or more points of a cluster become support vector a specific contour will be generated for the cluster. From a certain value of $q$, support vectors appear around each cluster.

\section[Geometric hashing based labeling]{Geometric hashing based labeling}\label{sec:ghbl}
In this section, we describe our mapping methodology to assign data points to clusters.

\subsection[2-d grid assumption]{2-d grid assumption}
In the previous method only support vectors guide processing to make contours but escape to know if a given point lies inside or outside the contour. Some methods such as describe in the foundation work by \cite{BenHur:2001}, and \cite{Yang:2002} work with an adjacency matrix defined as follows. Given two points of the data $x_{i}$ and $x_{j}$ and $\widehat{R}$ (the radius of the ball), the adjacency matrix $A$ such that:

\begin{eqnarray}
  \label{Aij}
   A^{i}_{j} = \left\{ \begin{array}{ll}
         1 \  \mathrm{if\ any\ point\ between\ }x_{i}\mathrm{\ and\ }x_{j}\mathrm{\ is\ such\ that\ }R(x_{k}) \leq \widehat{R};  \\ 
        0 \ \mathrm{ else}.\end{array} \right.
\end{eqnarray} 

Hence \cite{BenHur:2001} define a set of points between each pair and calculate if all the points belongs to the sphere or not, and so assign the pair of point to a cluster; In the second method \cite{Yang:2002} use a graph method to analyze the density areas of the graph defined by the adjacency matrix. We have compared our approach to these ones we call respectively in the following nearest-neighbors (NN) and minimum spanning tree (MST).
These methods are time-consuming and we imagined a method based on a geometric hashing function achieved with a grid surrounding data points in the attribute space. Basically according to the SVC method we only compute the radius for the points of the grid (that are hash keys) to build clusters, and as \cite{Datar:2004} we assume that almost closest points can be associated to a same hash. We use a nearest-neighbor method \cite{Cover:1967} to associate data points to their hash.

\subsection[Algorithm]{Algorithm}
The basic idea behind random projections is a class of hash functions that are locally sensitive; that is, if two points $(a, b)$ are close, they will have small $|p - q|$ values and they will hash to the same value with a high probability. If they are distant they collide with small probability. We have the following definitions.
Let $M$ be the size of the grid, and  fixed by the user.  A 2-dimension $M \times M$ grid is characterised with a step $s$. The step $s$ is defined according the minimal/maximal value of two first coordinates obtained by correspondence analysis (COA), $c1$ is the first coordinate, and $c2$ the second coordinate. We use the ade4 package of R-project to compute COA \cite{Dray:2007}. Let $g_{i}$ be a grid point.

\newtheorem{thm}{Theorem}[section]
\newtheorem{definition}[thm]{Definition}

\begin{definition} \textup{ ($s_{ck}$)  }  \\ 
\textup{ Let $s_{ck}$ be the scale of the grid from correspondance coordinates ck  }
\begin{equation}
  \label{s}
  s_{ck} = \frac{ ( max \left\{ g_{i}^{ck} \right\} ^{N}_{i = 1} - min \left\{ g_{i}^{ck}    \right\} ^{N}_{i = 1} ) } {M}, ck = \left\{ c1, c2 \right\}
\end{equation} 
\end{definition} 

We can define the set of grid points $g_{M}$ with each point $g_{i}$ by:

\begin{definition} \textup{ ($g_{G}$) }  \\ 
\textup{ Let $g_{M}$ be the set defined by: }
\begin{eqnarray}
  \label{gM}
  g_{G}(s_{ck}) = \left\{ g_{i}:dim(g_{i}) = 2\ \mathrm{and}\ (g_{i}^{ck} - g_{i + 1}^{ck}) = s_{ck},\ i\in\left[1:G\right] \right\}, ck = \left\{c1, c2\right\}
\end{eqnarray} 
\end{definition} 

For each point $g_{i}$ we can assess membership to clusters without specifying which one.

\begin{definition} \textup{ Let be $C = \left\{c_{j}\right\}$ the set of clusters, knowing radius R according Equation~\ref{R2} }   \\ 
\begin{eqnarray}
  \label{C}
  g_{i}\in C\ \mathrm{if} \left( \widehat{R} - R(g_{i}) \right) \geq 0 
\end{eqnarray} 
\end{definition} 

We now try to define clusters set with grid points:

\begin{definition} \textup{ (C) }  \\ 
\textup{We call C the set of clusters. A cluster consists of a grid point and all neighbouring grid points: }
\begin{eqnarray}
  \label{C_value}
  C = \{ c_{i}: \exists j\ g_{j} \in c_{i}\  \wedge\ g_{k} \in c_{i} \nonumber   \\ 
  \ \mathrm{if}\ g^{c1}_{k}\in\left[g^{c1}_{i} - 1, g^{c1}_{i} + 1 \right], g^{c2}_{k}\in\left[g^{c2}_{i} - 1, g^{c2}_{i} + 1 \right]\  \}
\end{eqnarray}
\end{definition}

Now we define the ball as the neighborhood of the hash key $(X, Y)$ from which it is assigned a specific cluster reference $c_{j}$ using a k-nearest neighbor threshold:

\begin{definition} \textup{ $B_k(X, Y)$ }  \\ 
\textup{Let $g_{G}$ the grid, C the set of clusters and $P (P_{c1}, P_{c2})$ a point with coordinates (X, Y) in the grid space GxG. Then the ball of P $B_k{X, Y}$ is defined by at least k neighbours belonging to a same cluster:}
\begin{eqnarray}
  \label{bk}
  B_k(X, Y) = \{ c_{j}: g_{i} \in c_{j} \nonumber   \\ 
  \wedge g^{c1}_{i}\in [X - 1, X + 1], g^{c2}_{i} \in [Y - 1, Y + 1] \wedge \#{i} \geq k \}
\end{eqnarray} 
\end{definition}

A family ${H = h:F \rightarrow G}$ is called locality-sensitive if, for any point $a$, the function $p(u)$ is defined as follows:

\begin{definition} \textup{ $p(u)$ }  \\ 
\begin{eqnarray}
  \label{pu}
  p(u) = Pr_{H}[ h(a) = h(b) = (X, Y):\left|a - b\right|\leq u, \nonumber   \\  
  E(a_{c1}) = E(b_{c1}) = X, E(a_{c2}) = E(b_{c2}) = Y ]
\end{eqnarray}
\end{definition}

$p(u)$ decreases in $u$. That is the probability of the collision of points $a$ and $b$ decreases with the distance between them.   \\ 
After defining a grid on data space, ClusterLabeling function achieves the first stage assigning a cluster number to each point of the grid.
The calculation of Lagrange coefficient gives the kernel matrix (MK). User settles the size of grid G, and MinMax value in data space can be computed. The main function (findSvcModel, described in next chapter) outputs a matrix called NumPoints linking each grid point to a cluster id. The radius Rc can be computed according algorithm shown in Table~\ref{table:Rc}.

\algsetup{indent=2em}
\newcommand{\Rc}{\ensuremath{\mbox{\sc Rc}}}

\begin{table}[H]
\begin{center}
\begin{tabular}{ll}
\begin{minipage}{6in}
\begin{algorithm}[H]
\begin{algorithmic}[]
\caption{ }

\medskip

\REQUIRE kernelmatrix MK, grid size G, MinMaxX min max value of  
          x value in data, MinMaxY min max value of y value in data.
\ENSURE NumPoints, a GxG vector for each grid point 
          and membership to a cluster id.
\medskip

\WHILE[we identify if a point belongs to a possible cluster] {each GxG Grid point P}
\STATE Associate\ x, y\ values\ to\ P\ from\ MinMaxX\ and\ MinMaxY
\STATE Calculate\ Radius\ Rp\ of\ P\ ,\ if\ Rp\ <=\ Rc\ ,\ give\ ball\ membership\ to\ P
\ENDWHILE

\medskip

\WHILE[we identify cluster id(s)] {each\ GxG\ Grid\ point\ P(i)}
    \WHILE {each\ P(k)\ around\ P(i)\ of\ one\ step}
       \IF {all\ points\ P(k)\ have\ no\  cluster\ membership}
           \STATE Create\ a\ new\ cluster\ vector\ CV\ with\ a\ new\ cluster\ id\ Cm
           \STATE Put\ CV\ in\ a\ list\ of\ cluster\ vector\ membership\ LCV
           \STATE Put\ P(i)\ to\ CV\ and\ associate\ Cm\ in\ NumPoints
       \ELSE 
           \STATE associate\ cluster\ id\ of\ P(k)\ to\ P(i)\ in\ NumPoints
       \ENDIF
    \ENDWHILE
\ENDWHILE
  
\medskip

\WHILE[we merge closed clusters] {each CV(i) in LCV}
    \WHILE {each\ other\ CV(j)\ in\ LCV\ !=\ CV(i)}
         \IF {CV(i)\ has\ distance\ of\ one\ step\ from\ CV(j)}
            \STATE Merge\ CV(i)\ and\ CV(j)
            \STATE Update\ NumPoints
         \ENDIF
    \ENDWHILE
\ENDWHILE
                   
\end{algorithmic}
\end{algorithm}
\end{minipage}
\end{tabular}
\end{center}
\caption{\label{table:Rc}
          Calculate Rc radius of the ball using MK.}
\end{table} 

Finally we can assign a cluster label for any point $x$ of the data set according the hash function and the corresponding ball value, defined in Equation~\ref{bk}. 

\begin{eqnarray}
  \label{fx}
  f(x) = c_{j}\ \mathrm{if}\ B_{k}( h(x) ) = { c_{j} }
\end{eqnarray} 

MatchGridPoint function, presented below, achieves the second stage; computation of $f(x)$ in Equation~\ref{fx}. It returns a vector we call ClassPoints associating a cluster id to each data point in the initial dataset (see Table~\ref{table:Matchgrid}).

\begin{table}[H]
\begin{center}
\begin{tabular}{ll}
\begin{minipage}{6in}
\begin{algorithm}[H]
\begin{algorithmic}[1]
\caption{ }

\medskip

\REQUIRE data matrix MD, grid size G, MinMaxX, MinMaxY, NumPoints, 
          neighbourhood of a data point k.
\ENSURE ClassPoints, a vector for data point and membership to a cluster id.

\medskip

  \FOR { each\ point\ D(i)\ in\ MD }
    \STATE Calculate\ Grid\ coordinate\ of\ any\ D(i)\ ,\ with\ MinMaxX,\ MinMaxY
  \ENDFOR

\medskip

  \FOR { each\ point\ D(i)\ in\ MD }
    \STATE Init\ a\ score\ vector\ SV(i)\ with\ dimension\ of\ cluster\ id(s)
    \FOR { each\ Grid\ Point\ P(j)\ in\ NumPoints }
       \IF { P(i)\ cluster\ id\ =\ k\ is\ found\ and\ distance\ between\ P(j)\ and\ D(i) = k }
         \STATE Increment\ SV(i)(k)
         \STATE Associate\ Max(SV)\ to\ Classpoint(i)
       \ENDIF
    \ENDFOR
  \ENDFOR
                   
\end{algorithmic}
\end{algorithm}
\end{minipage}
\end{tabular}
\end{center}
\caption{\label{table:Matchgrid}
          MatchGridPoint routine.}
\end{table} 

\subsection[Usage of the svcR package]{Usage of the svcR package}
Main function is the findSvcModel function. It computes a clustering model and returns it as an R object which is usable to other function for display and export.  Let call ret the return object, it covers some information about model parameters as the language coefficients (getlagrangeCoeff(ret)\$A attribute), the kernel matrix (getMatrixK(ret) attribute) and the cluster memberships (getClassPoints(ret) attribute). findSvcModel takes 10 arguments :

\begin{itemize}
\item data.frame means data.frame parameter in standard use  \\ or means data.frame in loadMat use \\ or means DatMat in Eval use, a matrix given as unic argument 
\item MetOpt,	optimization control parameter : optimStoch (stochastic way of optimization) or optimQuad (quadratic way of optimization)
\item MetLab,	labelling method: gridLabeling (grid labelling) or mstLabeling (mst labelling) or knnLabeling (knn labelling)
\item KernChoice,	kernel choice: KernLinear (Euclidian) or KernGaussian (RBF) or KernGaussianDist (Exponential) or KernDist (Matrix data as Kernel value)
\item Nu,	$nu$ parameter
\item q, $q$ parameter
\item k, $k$ nearest neigbours for grid
\item G, grid size
\item Cx,	$x$ component to display (1 for 1st attribute)
\item Cy,	$y$ component to display (2 for 2nd attribute)
\end{itemize}

If Cx and Cy are 0 the correspondent analysis is used. The data is given as first argument. The format is data.frame() (i.e. list) as the iris well known dataset. Some R libraries are required as quadprog \cite{Turlach:2007} for optimization, ade4 \cite{Dray:2007} and spdep \cite{Bivand:2010} for principal component analysis. This an exemple of usage in R : 
\\ 
\begin{verbatim}
R> library("svcR")
R> data("iris")
\end{verbatim}
\begin{verbatim}
R> retA <- findSvcModel(iris, MetOpt = "optimStoch", MetLab = "gridLabeling", 
+     KernChoice = "KernGaussian", Nu = 0.5, q = 40, K = 1, G = 5, 
+     Cx = 0, Cy = 0)
\end{verbatim}
\begin{verbatim}
R> plot(retA)
R> ExportClusters(retA, "iris")
R> findSvcModel.summary(retA)
\end{verbatim}

It means as data is the iris data frame. The Kernel choice is radial-based, parameters of SVC technique are $nu = 0.5$ and $q = 40$. Parameters for cluster labeling are $k = 1$ neighbor and grid size of $5 \times 5$ points. $Cx = Cy = 0$ means that first two principal components are used.  MetLab value means that geometric-hashing method is used. Plot function permits to visualize clusters. ExportClusters outputs clusters in a file with variables names. findSvcModel.summary displays size and number of clusters, and averaged attributes for each cluster. Some functions can help the user to navigate in clusters. ShowClusters(retA) returns all clusters ordered by their id (cluster 0 is a bag of variables not clusterable), GetClustersTerm(retA, term = "121") returns clusters in which "121" is a substring names of a member include in them, and GetClusterID(retA, Id = 1) returns the cluster with $Id = 1$.

\subsection[Toy example]{Toy example}\label{subsec:toy}
\begin{figure}
\centering
\begin{tabular}{cc}
\includegraphics[width=3in,height=2in]{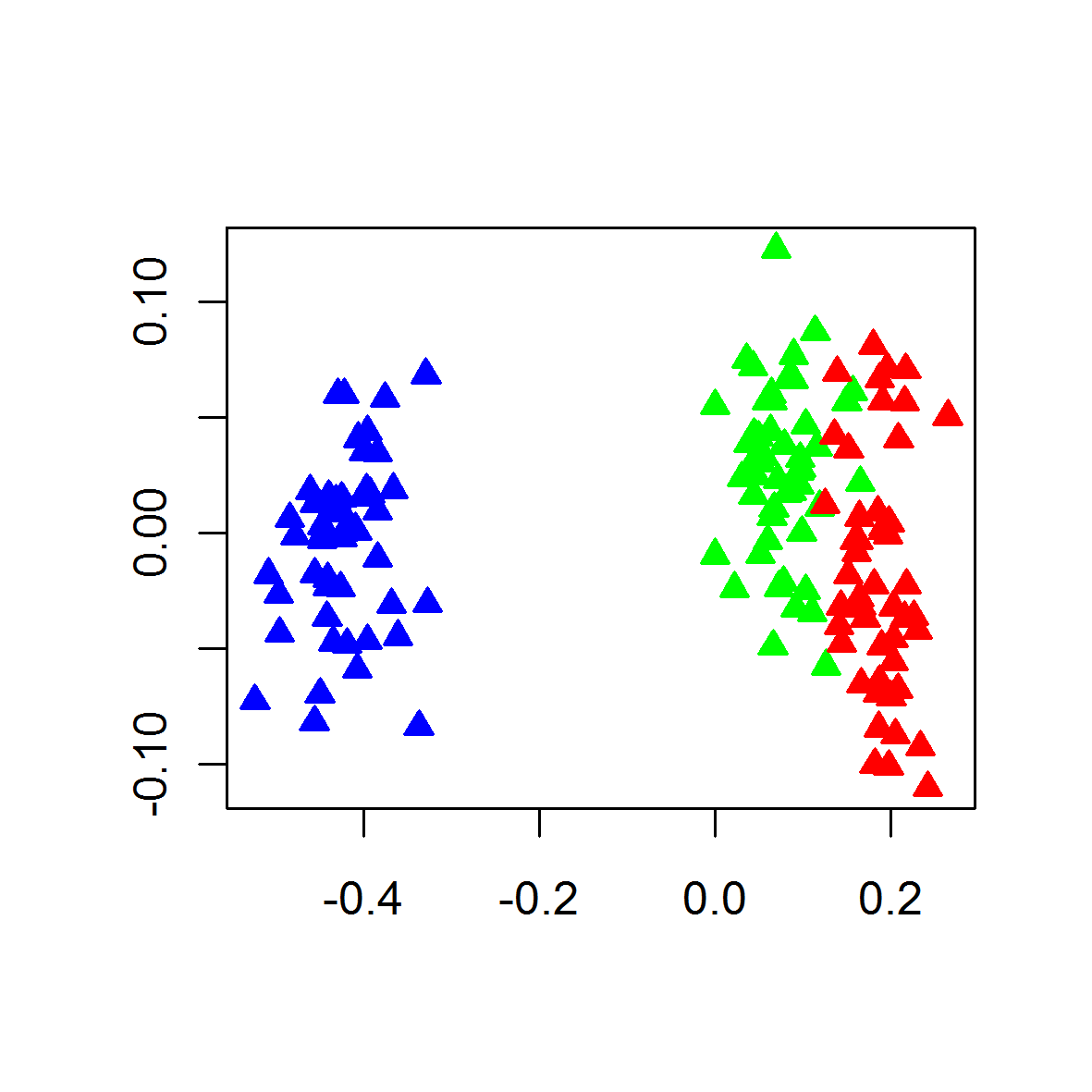} &
\includegraphics[width=3in,height=2in]{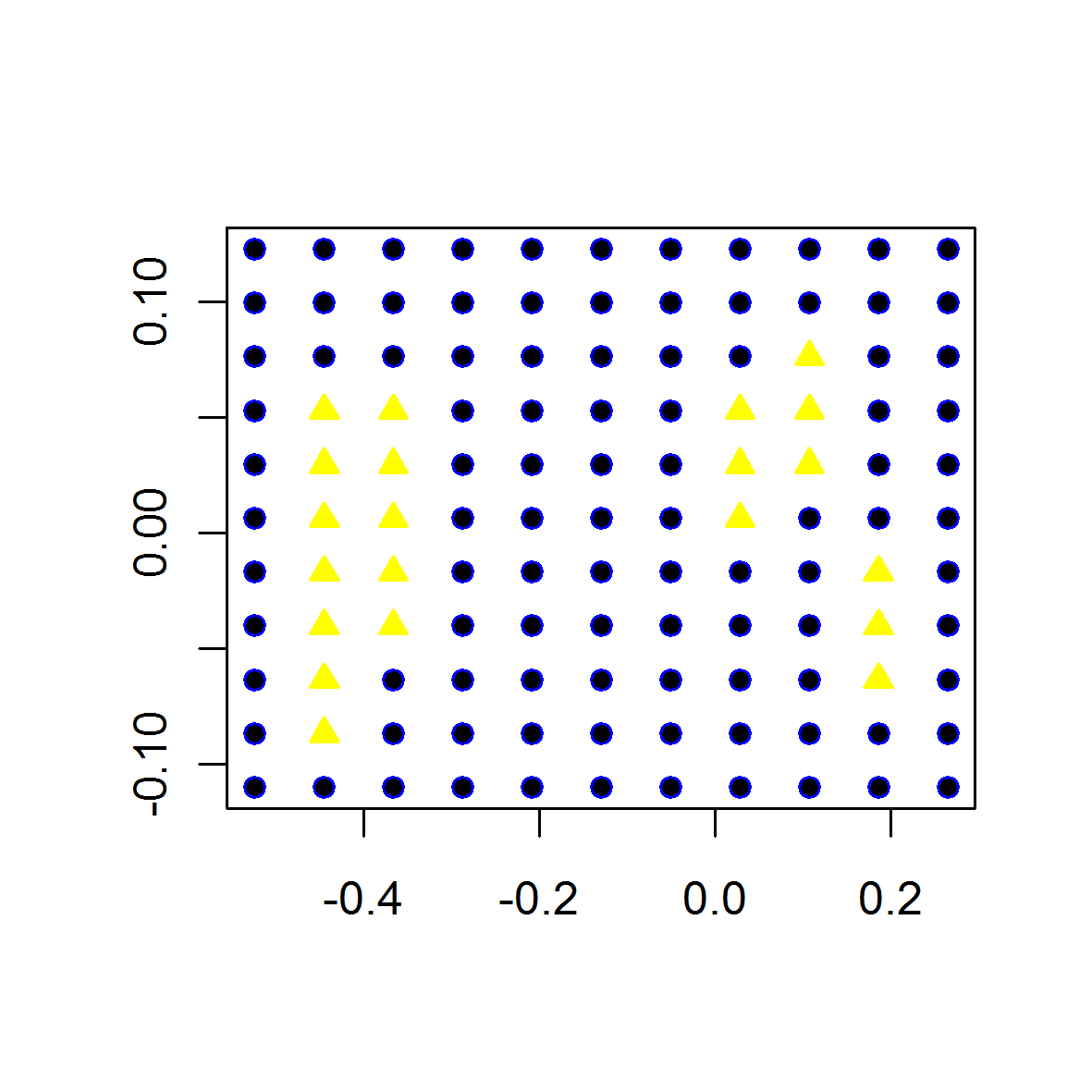} \\
\includegraphics[width=3in,height=2in]{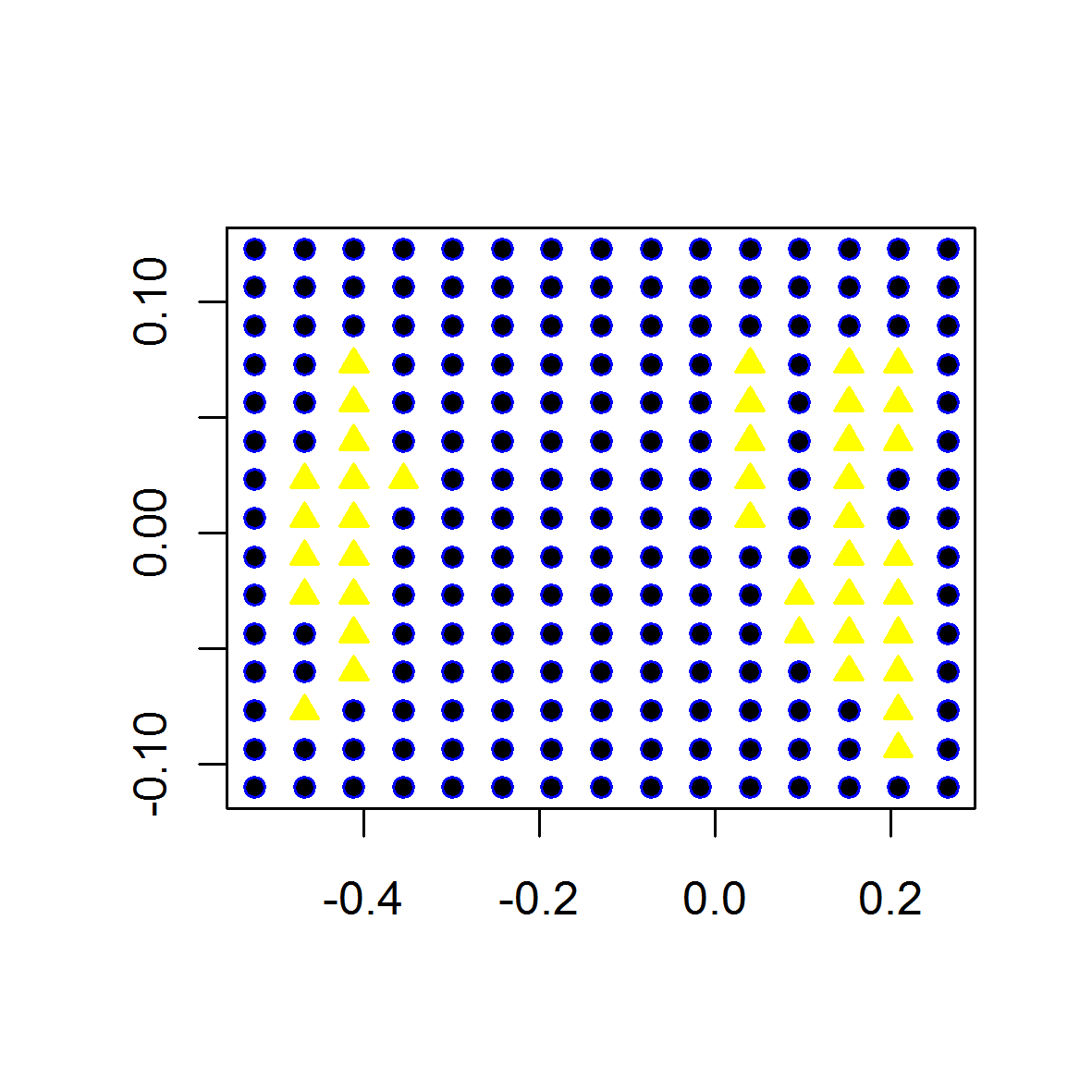} &
\includegraphics[width=3in,height=2in]{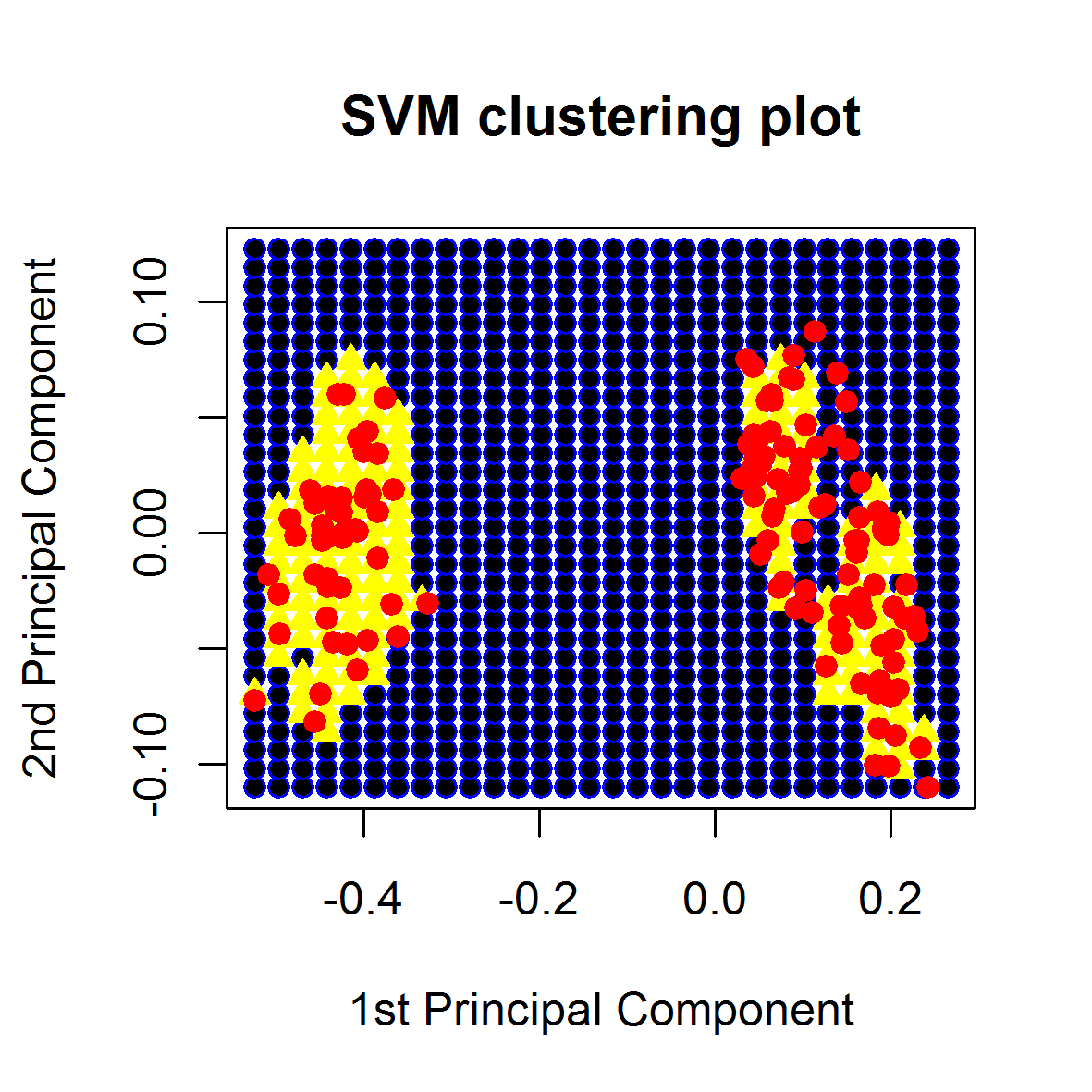}
\end{tabular}
\caption{2-D displays showing: data, clustered grid and data superimposed with clustered grid. Top left: Data plotted with COA $c1 = 1$, $c2 = 2$; Top middle $G = 11$, \#unclassified points is 17, \#missclassified points is 9; Top right: $G = 13$, \#unclassified points is 2, \#missclassified points is 7; Bottom: $G = 30$.}
\label{figure1}
\end{figure}

We used the famous Fisher's Iris data set. It contains 3 classes, 150 variables and 4 attributes. Our clustering extraction is largely based on the topology of points localized on a 2-D map. The dimensions of the maps are found by using a correspondence analysis and we kept the first two coordinates. The Iris data on these projection classes 2 and 3 are not well separated as it shown on Figure~\ref{figure1}. So the method can catch well class 1 and from time to time it occurs a "bridge" between class 2 and 3 that links them to form one cluster (Figure~\ref{figure1}).
The system is not very robust to force a so weak topological boundary. And so several iterations can force cluster 2 and cluster 3 to appear. For a grid size of $G = 13$, we obtain 50\% of success after a certain number of run executions.
\\

\begin{figure}
\centering{ \includegraphics[width=5in,height=3in]{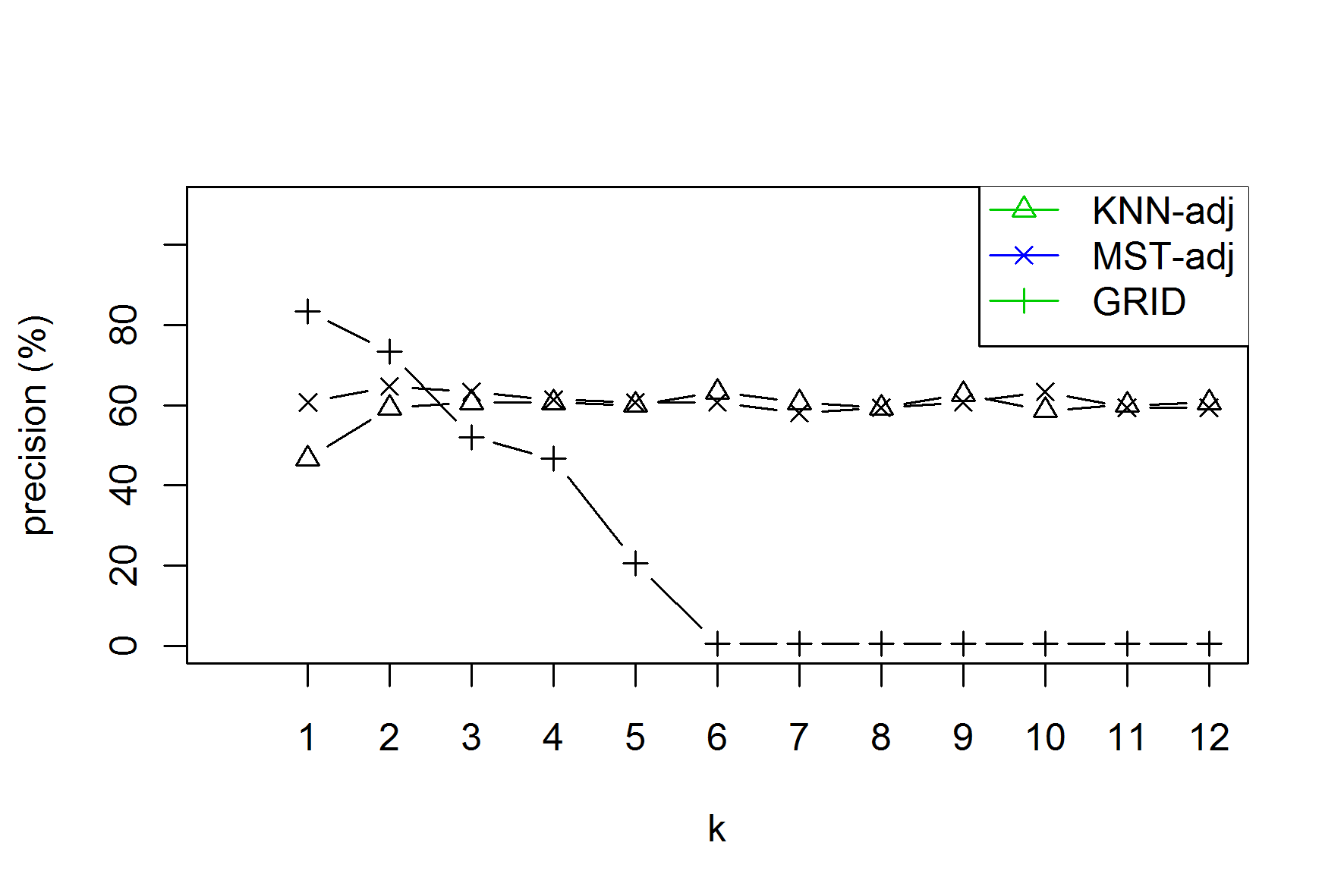} }
\caption{Clustering precision on iris dataset with parameters $Nu = 0.7$, $q = 1200$, $G = 13$ . K is the number of neighbours for GRID or KNN-adjacency or number of links for MST-adjacency approach.}
\label{figure2}
\end{figure}

The nearest neighbour parameter $k$ is used to find the closest cluster for a given data point. Low values such as $k = 1$ or $k = 2$ give same level of precision evaluation parameter to obtain 3 clusters. But this approach is not sufficient for good level of precision when the size of the Grid is high ($G > 25$) because the distance of peripheral data point is too far from their cluster. 
We compare precision of our approach ("GRID") with two other variants based on an adjacency matrix construction. The first variant makes the adjacency matrix with a minimum spanning tree ("MST-adj"), the second uses k-nearest neighbours ("KNN-adj"). All the three approaches have an order parameter we call k, that is the number of nearest neighbours for GRID and KNN-adj, and the number of links of a node in the tree for MST-adj. Mainly two clusters are captured by any approaches, and precision is computed by number of points of majority class included in cluster divided by the number of points (150). For GRID, precision when $k$ is small (between 1 and 3) is stable and competitive with other approaches (Figure~\ref{figure2}).

\begin{figure}
\centering { \includegraphics[width=5in,height=3in]{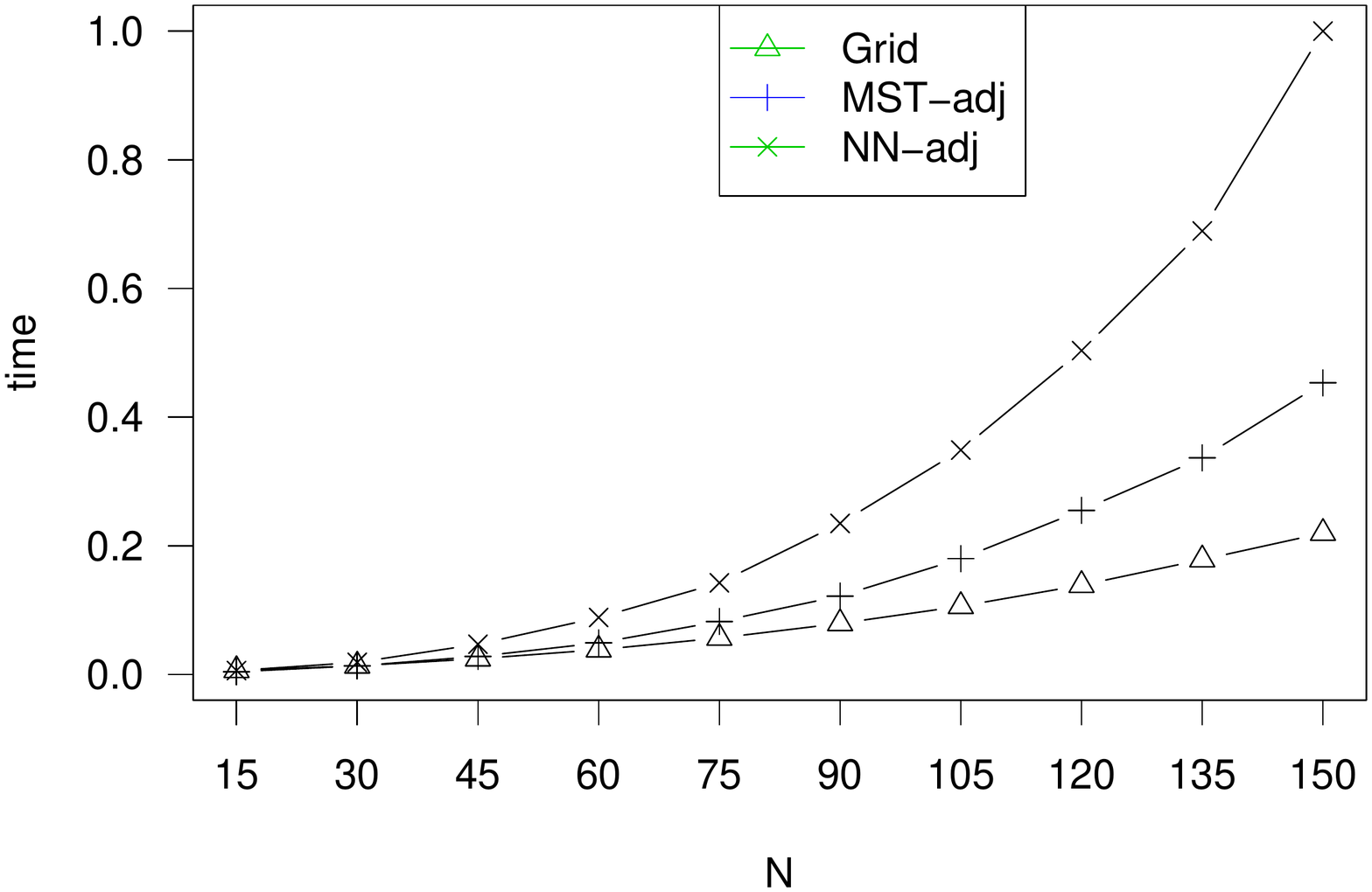} }
\includegraphics[width=5in,height=3in]{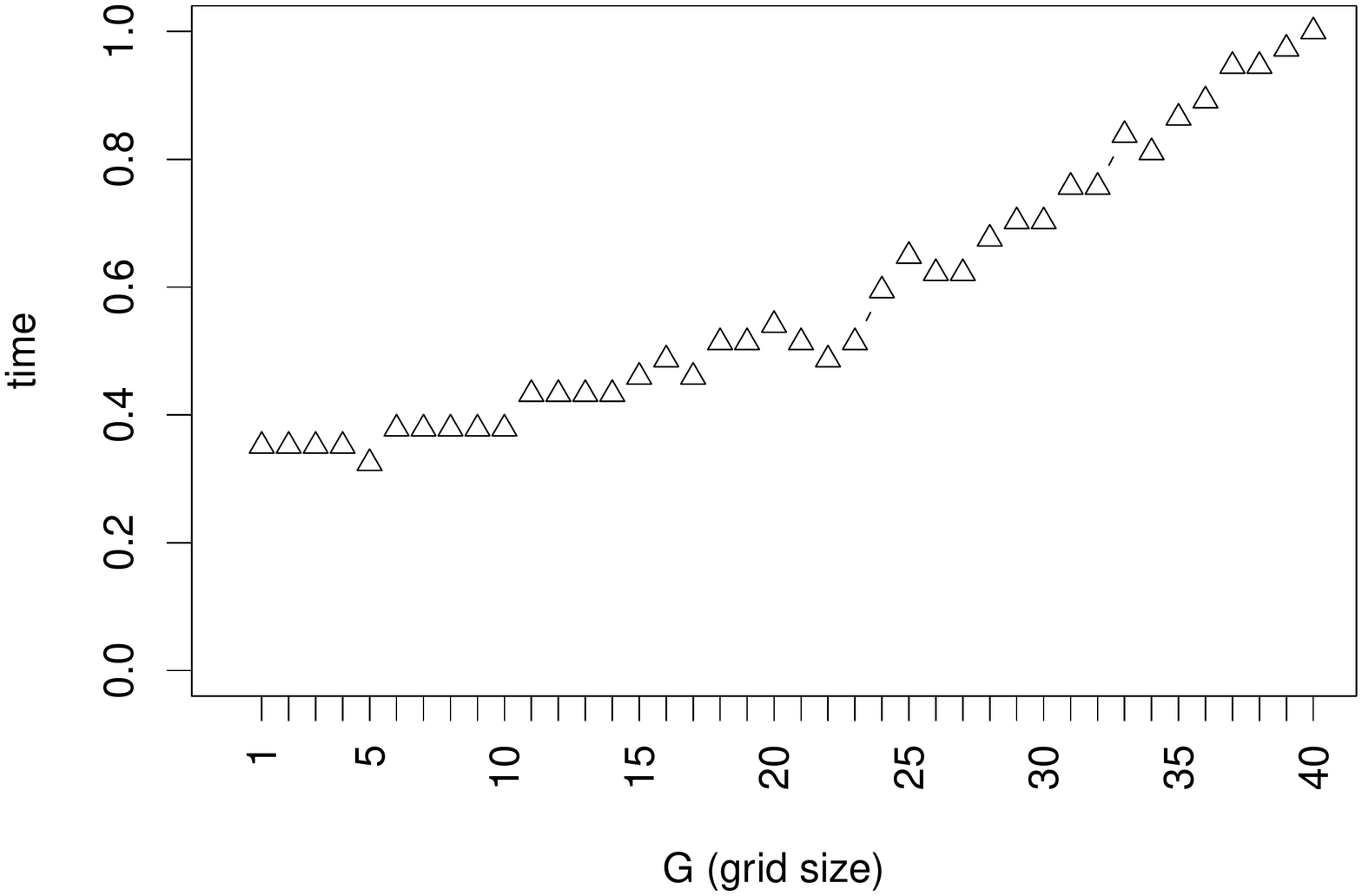}
\caption{Running time of svcR. At top, figure shows normalized speed of our Grid approach with MST-adj and KNN-adj according data size from 1 point to 150 points ( parameters of svcR are Nu = 0.7, q = 1200, G = 13). At down, figure shows normalized speed when grid size parameter of svcR increases from 5 to 40 points (Number of points = 150, Nu = 0.7, q = 1200).}
\label{figure3}
\end{figure}

A second stage of labelling using high-distance nearest neighbour should perform well at this size grid. But as we can see on Figure~\ref{figure3} (bottom) the running time for svcR is less interesting when $G > 30$ becomes higher, when $G$ is between 5 and 25 time run can increase by 25\% . We generally choose G in this range and the performance is not damaged compared to other approaches. If we look at Figure~\ref{figure3} (top) MST-adj is faster than KNN, and difference with svcR with 150 points is 2.05 times faster. Even with $G = 25$ we divides times by 1.65, hence we get back at least 40\% of time. In summary we can see that for $G = 13$ and a data size $N < 50$ for any method the run time is almost the same but increases very fast for the NN method when data size increases. Our approach becomes interesting for a much higher amount of data. For the whole Iris Data set our approach is two times faster but the run time depends on the grid size. We can see on Figure~\ref{figure3} that if $G > 25$ it becomes less competitive. If $G = 150$ time run is ten times than when $G = 40$, and twenty times than when $G = 20$. We used the quadprog package in R-Project for optimization.

\section[Representation of term sets and kernels]{Representation of term sets and kernels}\label{sec:rep}

In this section, we describe our good representation to classify term from a specific domain with an adapted kernel.

\subsection[Data]{Data, language models and domain knowledge}\label{subsec:data}

In the previous chapter we have shown that a radial base function can make a suitable clustering. But the data were made of a few attributes and not coming from natural language surrounded by sense ambiguities. We tried to make an attempt to classify terms coming from a specific domain: molecular and developmental biology. 

Our linguistic data set consists of 1,893 terms (linguistic phrases) manually extracted from an annotation of 1,471 documents (5,730 sentences) where annotated linguistic phrases describe temporal stages of biological development. The corpus itself has been build manually grabbed from Medline document database about spore coat formation and gene transcription specifically for Bacillus Subtilis species. We define some ideas about the language model studied in next chapter. Let suppose the following phrase "septal localisation of the division"; it will be supposed to be a term. From this term we can consider different sub-structures. "septal" and "localisation" are considered to be distinct words, and for instance "sept" is supposed to be a radical i.e a sequence of character which can be found in other words. "septal localisation" is considered as a bigram, i.e. a sequence of two words. "localisation of the" is considered as a trigram, i.e. a sequence of three words. 

Textual corpus we used describes biological knowledge and especially a well known biological model called sporulation. This biological process is activated by a microorganism to be resistant in an environment with starvation. The bacterial is transformed into a resistant sphere with mininum needs and activity. In information extraction from texts gene network reconstruction is a quite interesting field to understand how a gene network is activated. Temporal and spatial information are complementary information useful to understand when gene interactions occured. A well studied biological process as sporulution can be a reference model with both interest:
\begin{itemize}
	\item Gathering enough molecular information about gene-gene interactions in texts since ten years;
	\item Being a well described biological model across different stages.
\end{itemize}
Six main stages describe the sporulation process. At the beginning of the process a frontier called the septum is created and at the last stage an engulfment is created to leave out the bacterial spore. The 1,893 terms have been also classified manually into the 6 biological stages. An average amount of 600 terms can cover a given stage. The problem is related to morphological and fuzzy description of language. Where a strict formal description should used for instance "stage II" concerning the second stage of biological development, an expert could use "during the first stages of sporulation" or "at the onset of sporulation" or "at stage I-II" or "after septum formation" ...etc. Moreover complexity of description, we can imagine insofar because 600 terms per class on to only 1,893, is that lots of terms are not exclusive to one stage (i.e. one class). Lots of expressions can designate a stage and often several stages at the same time. 

Why do a clustering method such as SVC could be of interest ? We observed that: 
\begin{enumerate}
	\item Most of terms describing occurrence of a gene activation/inhibation/regulation are not expressed in a simple regular way such as "at stage 2" or "at stage 3". But terminology of temporal knowledge has a variable expressivity;
	\item Lots of terms are not exclusive to a stage.
\end{enumerate}

In such usage context, the svcR technique could help an expert to build rules about expressions to get equivalence between a set of expressions and a mapping of rule with a specific class. We decided to compare which language model can bring benefit for term description and for each language model which kernel can be also more relevant. We had manually selected a list simple morphological radicals (11 tokens), word bigrams (a restricted sample of 500 on to 1,477) and word trigrams (a restricted sample of 500 on 2,179) from the whole set of terms. Figure~\ref{figure4} gives a sample of some linguistic expressions. In our clustering experiments we first made a sample of 98 terms and 4 classes, similar in size with iris data (Section~\ref{subsec:toy}).

\begin{figure}
\centering
\renewcommand{\arraystretch}{1.2}
{\scriptsize
\begin{tabularx}{15.5cm}{|>{\centering\hsize=0.4\hsize\arraybackslash}X|>{\centering\hsize=0.4\hsize\arraybackslash}X|>{\centering\arraybackslash}X|>{\centering\hsize=0.8\hsize\arraybackslash}X|>{\centering\hsize=1.\hsize\arraybackslash}X|>{\centering\arraybackslash}X|>{\centering\hsize=0.5\hsize\arraybackslash}X|}
   \hline
   \multicolumn{4}{|c|}{ \textbf{Terms (TM)} } & \textbf{Radicals (RD) } & \textbf{Bigrams (BG)} & \textbf{Trigrams (TG)} \\
   \hline
   \textbf{class 1} & \textbf{class 2} & \textbf{class 3} & \textbf{class 4} &  &  &  \\	 	 	 
   \hline
   insertion into the septum & prespore development & cortex layer , synthesized between , the forespore inner and , outer membranes & coat, encases the spore & init & cell specific & the mother cell \\
   \hline
   integrity of the septum & prespore gene expression & cortex peptidoglycan in spores & coats & sept & spore coat & mother cell specific \\
   \hline
   septal compartment & prespore programme of gene expression & cortex structure & coats of wild-type spores & prespore & during sporulation & in the mother \\
   \hline
   septal localization of the division & prespore-like cells & cortexless spores & compartment & endospore & and sporulation & mother cell compartment \\
   \hline
   septal peptidoglycan during cell division & prespore-specific & cortical or vegetative peptidoglycan synthesis & compartment-specific & engulfment & of sporulation & growth and sporulation \\
   \hline
\end{tabularx} } 
\caption{Samples of terminological data sets.}
\label{figure4}
\end{figure}

After viewing which language model (term-radical, term-term, term-bigram, term-trigram) and which kernel are enough efficient, we apply the language model and the kernel to the whole set of 1893 terms.

\subsection[Kernels]{Kernels}

As terms (that are strings), intrinsically and without textual context, can be statistically compared pairwise (in a Levenshtein way) or using a bag-of-words (in the Jaccard way) we compared these approaches, in addition to robustness due to randomized non null value in the Jaccard case. 
The Levenshtein distance is an editing distance based on the cost to transform a string into another \cite{Levenshtein:1966}. Assume $a$ and $b$ being two strings. Let $a_{i}$ be the sub-string consisting of the first $i$ symbols of string a where $0\leq i\leq\|a\|$ and $b_{j}$ be the sub-string consisting of the first $j$ symbols, iteratively we obtain the Levenshtein distance at position $i$ and $j$:

\begin{eqnarray}
  \label{Dij}
  D_{i, j} = min(D_{i - 1, j} + w^{I}, D_{i - 1, j - 1} + w^{S}, D_{i, j - 1} + w^{D})
\end{eqnarray} 

where $w^{I}$, $w^{D}$ and $w^{S}$ are weights of insertion, deletion and substitution on operations respectivaly and $D_{0, 0} = 0$
Finally $D_{a, b}$ represents the weighted Levenshtein distance. From its expression we define the Levenshtein radial base kernel:

\begin{eqnarray}
  \label{LRB} 
  LRB(x_{1}, x_{2}) = e^{ -q.\left\| \sum^{N}_{i = 1}(D_{1i} - D_{2i})\right\|^{2}}
\end{eqnarray} 

We also define a kernel using only the component of Levenshtein distance between a pair of terms:

\begin{eqnarray}
  \label{RBL}
  RBL(x_{1}, x_{2}) = e^{ -q.\left\|D_{12}\right\|}
\end{eqnarray} 

Equation~\ref{LRB} and Equation~\ref{RBL} are a composition of a semi-positive definite kernel (the radial base function) so the final kernels are also semi-definite positive.
The Jaccard index is a similarity index \cite{Jaccard:1901} that is useful to assess the similarity between two objects computed only knowing the set of their attributes, and not the whole set of attributes being often huge and not describing the given objects. Its expression is the following knowing that a string $s_{1}$ is composed with tokens {$s_{10}$,..., $s_{1m}$} and string $s_{2}$ is composed with tokens {$s_{20}$,..., $s_{2k}$}:

\begin{eqnarray}
  \label{J12}
  J_{12} = J(S_{1}, S_{2}) = \frac{ \left| \left\{ S_{10},..., S_{1m} \right\} \cap \left\{ S_{20},..., S_{2k} \right\} \right| }{ \left| \left\{ S_{10},..., S_{1m} \right\} \cup \left\{ S_{20},..., S_{2k} \right\} \right| }
\end{eqnarray} 

Hence we define a Jaccard-radial base kernel (JRB) according vector defined with Jaccard index with other terms (the data matrix is symmetric):

\begin{eqnarray}
  \label{JRB} 
  JRB(x_{1}, x_{2}) = e^{ -q.\left\| \sum^{N}_{i = 1}(J_{1i} - J_{2i})\right\|^{2}}
\end{eqnarray} 

We also define a kernel using only the component of Jaccard index between a pair of terms:

\begin{eqnarray}
  \label{RBJ} 
  RBJ(x_{1}, x_{2}) = e^{ -q.\left\|J_{12}\right\|}
\end{eqnarray} 

Equation~\ref{JRB} and Equation~\ref{RBJ} are a composition of a semi-positive definite kernel (the radial base function) so the final kernels are also semi-definite positive.
\cite{Xu:2004} and \cite{Bilenko:2002} have been respectively adapted a kernel approach with a Levenshtein and a Jaccard similarity coefficient and proved their robustness though their classical simplicity. In our data representation we have used four Kernels: the Levenshtein-radial base (LRB), the radial base-Levenshtein (RBL), the Jaccard-radial base (JRB) and the radial base-Jaccard (RBJ). We also have introduced noise in the data matrix such that if the Jaccard coefficient gives 0 we assign a random non null value to the data matrix component. We call this fuzziness, Jaccard+.
The vector approach using such distance and index heuristics in natural language processing sets the representation of description by sets of words but property of such sets can be modulated. For instance co-frequency in textual context (with left and right collocations) (lexical-based similarity), or string inclusion between two terms (dictionary-based similarity), or ontological nodes shared between two terms (conceptual-based similarity). We focused on the second way and we compared several cases of dictionary to build the matrix of similarity. As variables to classify of course we used the sets of terms, and as attributes the set of radicals (RD), the sets of terms itself (TM), the sets of bigrams (BG) and the set of trigrams (TG).

\subsection[Results]{Results}

\begin{figure}
\centering { \includegraphics[width=400pt]{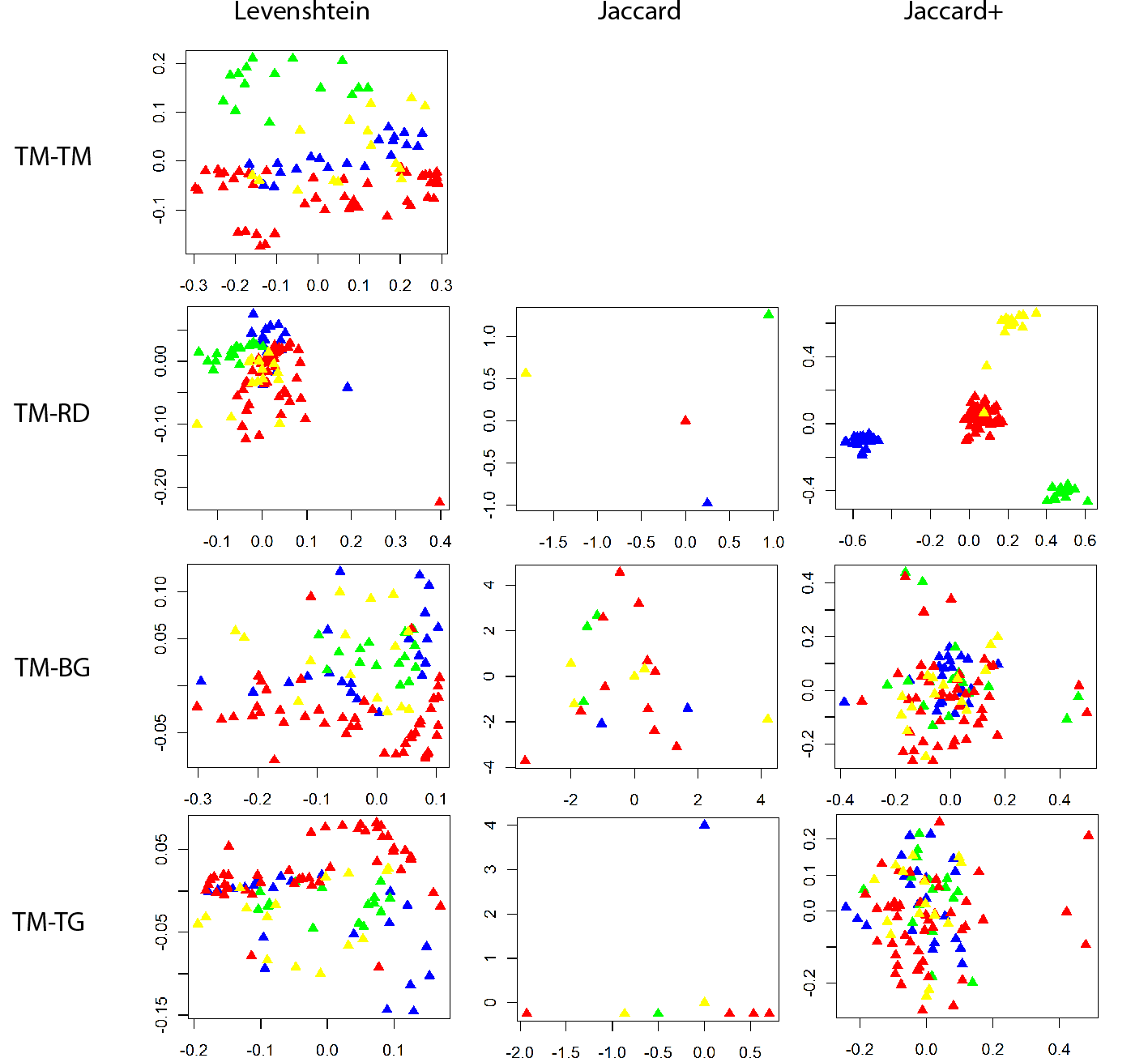} }
\caption{In this table each display is dependant upon features describing linguistic phrases (TM or terms) i.e. with terms themselves (TM-TM), with radicals (TM-RD), with bigrams (TM-BG) or with trigrams (TM-TG), secondly results depends upon kernel used for clustering Levenshtein, Jaccard or Jaccard with artificial noise. Displays represent data classes in green, red, blue, yellow colors and in 2-d maps of the COA components.}
\label{figure5}
\end{figure}

\begin{figure}
\centering { \includegraphics[width=400pt]{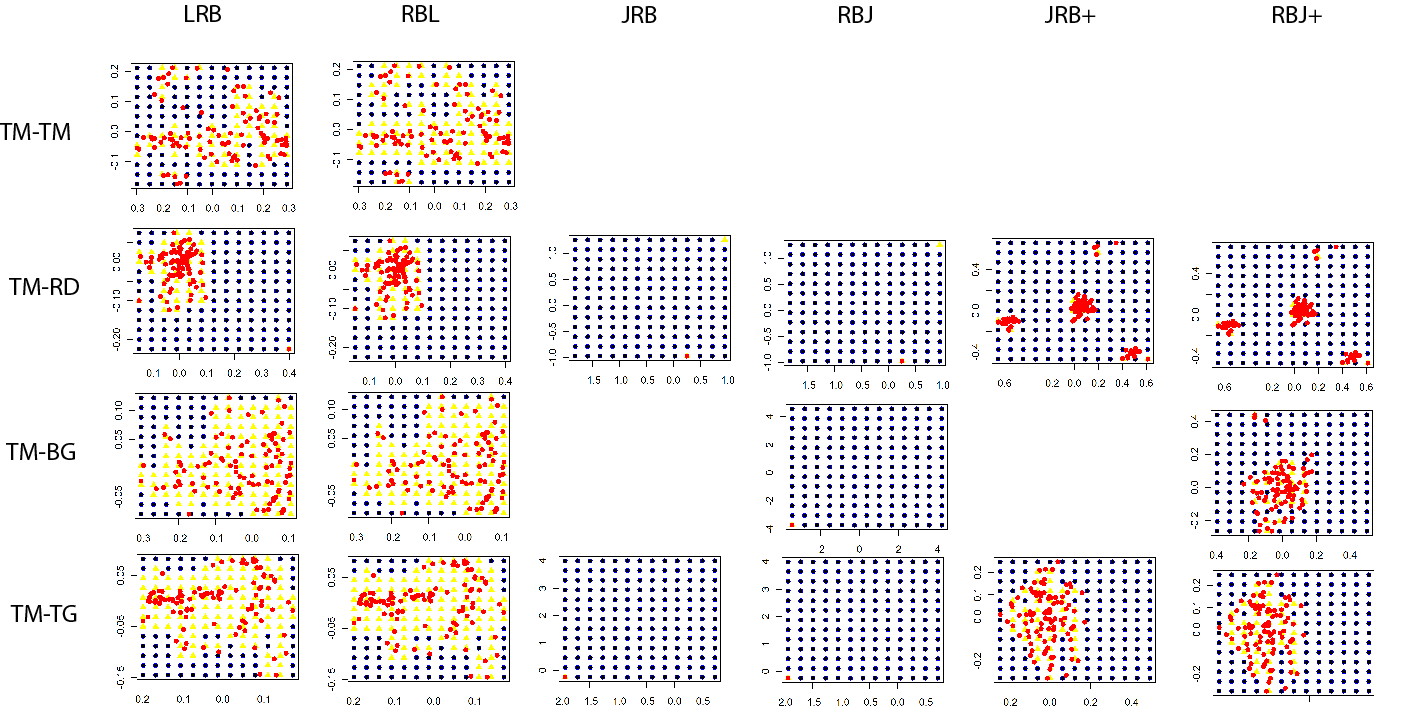} }
\caption{Clustering with SVC-geometric hashing ($Nu = 0.9$, $G = 13$, $q$ chosen at best between 1 and 10000).}
\label{figure6}
\end{figure}

As we see in the Section~\ref{sec:svc} for the presentation of the SVC method coupled to our geometric approach for cluster extraction, if no clear geometric separation in data occurred on the 2-D map of correspondence analysis coordinates, the method is unsuccessful. Figure~\ref{figure5} shows different plots of the different cross between data attributes and distances. We see on these maps that TM-TM Levenshtein, TM-RD-Jaccard and TM-RD Jaccard+ can produced interesting maps for SVC application.
Thus on each set of the data matrix we applied the cluster extraction to compare the efficiency of class retrieval. Figure~\ref{figure6} shows performance of the method. The Jaccard kernel gives best results with a good separation and extraction of classes. And the variant set introducing random noise in the matrix still becomes successful with 2 misclassified items on 98.
\\  

\begin{figure}
\centering { \includegraphics[width=400pt]{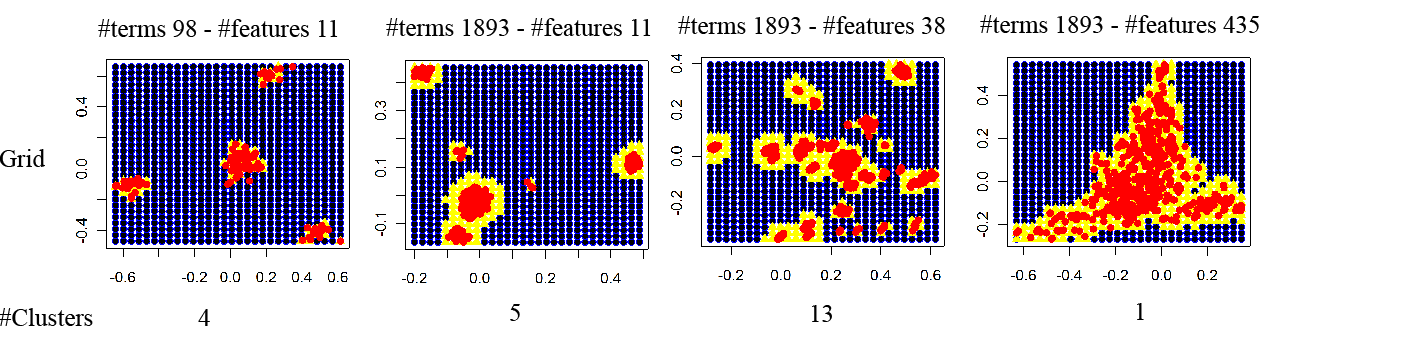} }
\caption{Clustering with SVC-geometric hashing ($Nu = 1$, $G = 30$, $q = 2000$, JRB+); Each column means different number of terms and number of features, datasets sizes increase from left to right. Below are the number of clusters extracted with svcR. Yellow color represent clusters, red is data color for major class of a cluster and green is data color but not belonging to major class of a cluster.}
\label{figure7}
\end{figure}

Now we adopt the best obtained clustering setting that is a term-radical matrix (language model) and Jaccard Radial base kernel. Now to study scalability efficiency we expand amount of terms and radicals taken into account for Jaccard distance computation. Independently of clusters purity (class homogeneity), impact of features (radicals) is a warranty to make a good separation between similar terms. We do not forget that support vector machine is a non linear method which is efficient only if data are separable. Hence recall that role of features is to make similarity clue between terms, role of Jaccard index is to capture similarity, role of 2D component analysis is to capture main features that make separation between data, and finally role of support vector clustering is to capture bounds of cluster thanks to their geometric separation. Figure~\ref{figure7} shows that too many features do not make separation of data (attribute DName is changing for each four data sets):
\\  

But too few features make too few set of clusters. A medium set of features can lead to a good number of clusters. In our case 38 features describing structure of terms induce 15 clusters easily distinguishable visually. Recall the set of terms is made of 1893 terms describing 6 stages of sporulation process as we mentioned in Section~\ref{subsec:data}.

Lots of terms belong to several stages (in the sense of classes). Even typical string token relevant of a class can belong to different stages. It is mainly due to biological stage results from microscopy studies, so visual patterns and often a co-occurrence of patterns can be simultaneously typical of a stage but individually we can observe a pattern occurs during several stages as \textit{mother cell} and \textit{compartmentalization} (beginning at stage 2 and staying at stages 3, 4, 5, 6), \textit{engulfment} (beginning at stage 3 and staying at stages 4, 5, 6), \textit{septum} (beginning at stage 1 and staying at stages 2, 3, 4). This property of cross-membership is hard enough to compute as a mapping between a specific term to a unique class. In our results (Figure~\ref{figure7}) getting more clusters (15) than classes (6) induces that terms can be misclassified (green points) but make a variety of specific clusters from which we expect they capture patterns association that should be used to define rule of an automaton. For instance among the 15 clusters, a specific one gives the following members :   \\ 
\centerline{    \textit{ compartmentalized activation , compartment-specific activation } } 
We can imagine a rule associated to stage 2, 3, 4 and 5 : "compartment\* activation". Another one gives the following members : \newline
\centerline{ \textit{ slow postseptational, prespore-specific SpoIIIE synthesis, endospore coat, }}
\centerline{ \textit{ endospore coat assembly, endospore coat component, forespore coat, from the endospore coat, }}
\centerline{ \textit{ cortex and/or coat assembly, spore coat and cortex.}}
We can imagine a rule associated to stages 3, 4, 5 and 6 : "endospore coat", "coat \* cortex", "cortex \* coat".
From these clusters of terms coat, cort, prespore, endospore, postseptational, forespore are in the sets of features. In this process of lexical rule definition the user plays an important role in such way a cluster do not give information directly exportable as an automatically defined rule. Visualization of clusters by an expert leads to identification of patterns association to include in lexical rules. Especially by the fact that elements taken into account are features and knowledge about features is required to say that these rules will be applicable to a set of classes (biological stages). The methodology makes us to understand, but it is not a discovery, that clustering mixes several components of different categories. Nevertheless it can be efficient to identify relevant features to identify as a lexical pattern to build rules for information extraction, in our case information extraction of a biomolecular process described linguistically and formally by several stages (i.e. a scenario in the domain of biology).

\section[Comparison with other techniques]{Comparison with other techniques}\label{sec:comp}

In this section, we discuss behavior of concurrent clustering methods, existing kernels and interpretation of SVC clustering capacity. Below a simple general R utility function, gets outputs of used R clustering functions (k-means, svcR, hierarchical) and exploits a data property that is insertion of the class number in each term (as "4   coat protein" meaning "coat protein" belongs to class 4). Hence using grep function it is very easy to find the repartition of classes over clusters :

\begin{verbatim}
TabEval <- function(Dat) {		
M <- matrix(nrow = (max(Dat[]) + 1), ncol = 8, 0)	
for( k in 1:max(Dat[]) ){			
	Stat <- c()					
	Size <- length(Dat[ Dat == k])		
	for(i in 1:6) {				
		GR   <- grep(i, names(Dat[ Dat == k]) )	
		Stat <- c(Stat, 100 * length(GR) / Size )	
	}							
	Stat= c(Stat, 0, Size )			
	M[k,] <- Stat				
}						
Stat <- c()					
for(i in 1:6) {				
	Size <- length(Dat[])			
	GR <- grep(i, names(Dat[]) )		
	Stat <- c(Stat, 100 * length(GR) / Size )	
}						
Stat <- c(Stat, 0, Size )				
M[nrow(M), ] <- Stat				
print(M, digits = 1)			
\end{verbatim}

\subsection[Classical clustering]{Classical clustering}

\begin{figure}
\centering{ \includegraphics[width=1.8in,height=1.7in]{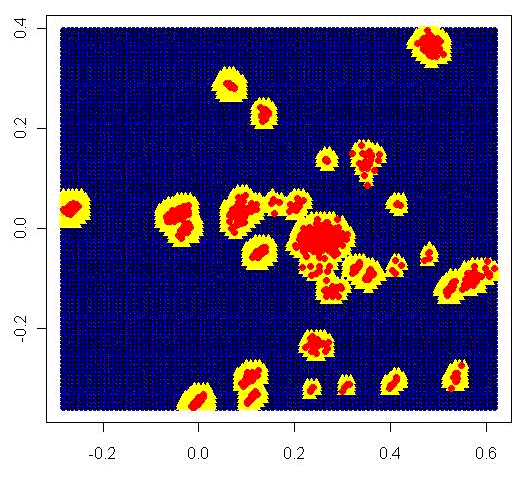}  }
\includegraphics[width=1.8in,height=1.7in]{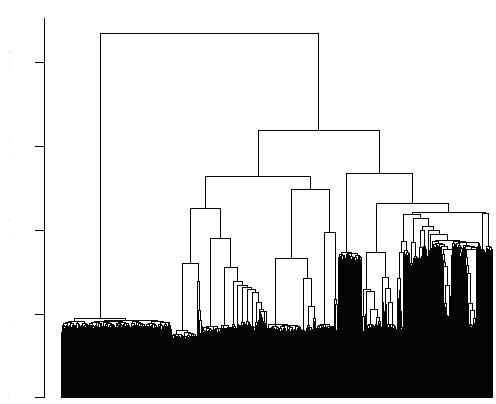} 
\includegraphics[width=2in,height=2in]{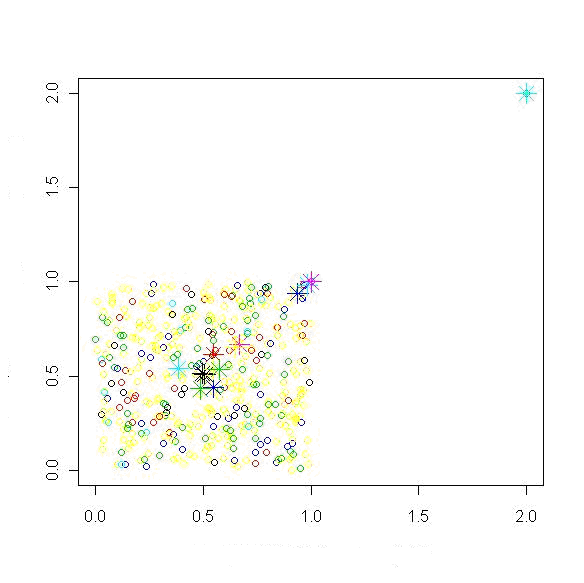} 
\caption{Clustering with SVC-geometric hashing (left), hierarchical agglomerative clustering (centre), k-means (right); Data are 1893 terms with 6 classes and 37 features using a Jaccard-radial base kernel.}
\label{figure8}
\end{figure}

Algorithms such as k-means (KM) and hierarchical clustering (HC) are widespread poor knowledge techniques using metrics to find automatically clusters in any kind of data. Figure~\ref{figure8} shows graphically how such clusters could be represented.

About svcR and KM, 2-dimensional coordinates come from component analysis. On the KM map only centroids represent clusters (as star plotting characters). HC (Figure~\ref{figure8}, center) displays a dendrogram where branches mean clustered points and require a cut-off at a level of the tree to catch clusters. In R, we used kmeans function from stats package \cite{R-project:2010}, and hclusterpar function from amap package \cite{Lucas:2007}.

As Data contains 6 classes and svcR approaches with JRB kernel induces extraction of 17 clusters we settle 30 clusters extraction as settings for both KM and HC function. Figure~\ref{figure9} shows the content of clusters and class distribution for each approach (hierarchical, k-means and SVC). The right column of each result set means the size of each cluster. The last line means distribution over classes from the whole dataset as baseline (it means that 12\% of terms belong to class 1, 19\% to class 2, 20\% to class 3, 20\% to class 4, 16\% to class 4, 12\% to class 6 and size of set is 1893 terms). First we can observe that distribution profile in cluster size is similar between HC and svcR. Secondly, looking at over-representation of classes over clusters HC and KM do not achieve better discrimination of terms across the 6 classes some clusters are better over. Language ambiguities seem to be a real bottleneck for all methods when usage is based on a Jaccard-Radial Kernel. But what happens when string kernels are used ?

\subsection[String kernels]{String kernels}

\cite{Lodhi:2002} and \cite{Moschitti:2009} promoted kernel strings to get semantic knowledge from texts. The string kernels calculate similarities between two strings by matching the common substring in the strings. A standard string kernel is the constant one (SK-constant) and assess similarities even is characters are matching in any order, and higher is the return value when order is respected and size of matching is bigger. Exact matching of $n$ characters is called spectrum kernel (SK-spectrum) \cite{Teo:2006}. For instance let suppose a string of 29 characters and estimate value of a string with itself, SK-constant return 3165, SK-spectrum return 430. 

\begin{figure}
\centering
\renewcommand{\arraystretch}{1.2}
{\scriptsize
\begin{tabularx}{10.17cm}{|>{\centering\hsize=1.3\hsize\arraybackslash}X|>{\centering\hsize=1.3\hsize\arraybackslash}X|>{\centering\hsize=1.3\hsize\arraybackslash}X|>{\centering\hsize=2\hsize\arraybackslash}X|>{\centering\hsize=1.3\hsize\arraybackslash}X|>{\centering\hsize=1.3\hsize\arraybackslash}X|>{\centering\hsize=1.3\hsize\arraybackslash}X|>{\centering\hsize=2\hsize\arraybackslash}X|>{\centering\hsize=1.3\hsize\arraybackslash}X|>{\centering\hsize=1.3\hsize\arraybackslash}X|>{\centering\hsize=1.3\hsize\arraybackslash}X|>{\centering\hsize=2\hsize\arraybackslash}X|}
   \hline
   \multicolumn{4}{|c|}{ \textbf{HC} } & \multicolumn{4}{|c|}{ \textbf{KM} } & \multicolumn{4}{|c|}{ \textbf{svcR} }  \\
   \hline
   \textbf{C1} & \textbf{C2} & \textbf{C3}  & \textbf{\#} & \textbf{C1} & \textbf{C2} & \textbf{C3}   & \textbf{\#} & \textbf{C1} & \textbf{C2} & \textbf{C3}  & \textbf{\#}     \\
   \hline
   0.17 & 0.17 & 0.16   & 481 & 0.23 & 0.27 & 0.16   & 153 & 0.17 & 0.17 & 0.16   & 481    \\
   \hline
   0.22 & 0.24 & 0.16   & 152 & 0.04 & 0.18 & 0.24   & 49 & 0.11 & 0.17 & 0.23   & 283    \\
   \hline
   0.14 & 0.16 & 0.25 &   199 & 0.22 & 0.16 & 0.20   & 64 & 0.03 & 0.25 & 0.22   & 156    \\
   \hline
   0.27 & 0.27 & 0   & 15 & 0.03 & 0.22 & 0.22   & 103 & 0.04 & 0.20 & 0.21   & 113    \\
   \hline
   0.17 & 0.17 & 0.17   & 78 & 0.09 & 0.27 & 0.18   & 11 & 0.15 & 0.31 & 0.26   & 81    \\
   \hline
   0.30 & 0.49 & 0.07   & 61 & 0 & 0.10 & 0.31   & 29 & 0.06 & 0.21 & 0.19   & 63    \\
   \hline
   0.13 & 0.13 & 0.20   & 15 & 0.17 & 0.11 & 0.19   & 54 & 0.17 & 0.17 & 0.17   & 78       \\
   \hline
   0.12 & 0.08 & 0.27   & 26 & 0.04 & 0.28 & 0.20   & 50 & 0 & 0.28 & 0.39   & 18    \\
   \hline
   0.04 & 0.24 & 0.21   & 219   & 52 & 0 & 0 & 44 & 0 & 0 & 0.25   & 4    \\
   \hline
   0.25 & 0 & 0.25   & 4 & 0.17 & 0.24 & 17   & 41 & 0.08 & 0.75 & 0.17   & 12    \\
   \hline
   &  &  &  &  &  &  &  &  & &  &    \\
   \hline
   \textit{0.12} & \textit{0.19} & \textit{0.20}   & \textit{1893} & \textit{0.12} & \textit{0.19} & \textit{0.20}   & \textit{1893} &  \textit{0.12} & \textit{0.19} & \textit{0.20} &   \textit{1893} \\
   \hline
\end{tabularx} } 
\caption{Class distribution over clusters resulting from HC (left 4 columns), KM (centre 4 columns) and svcR (right 4 columns). Only three first classes are displayed (over 6) i.e. C1 to C3. These classes are hand-made built and each term is tagged in the data matrix with one of these classes. After clustering, we collect class membership of terms for a given cluster. In the table, each line is the distribution of a given cluster (different over each method HC, KM or svcR). Each line shows the weigth in percent for each class in the cluster. The last line represent a baseline, showing how should be the weigth of a class if all terms should form a unic cluster. Hence for instance class 1 represents 12\% of the set of terms. On each line we dispaly also (column \#) the number of terms belonging to the cluster.}
\label{figure9}
\end{figure}

\begin{figure}
\centering{ \includegraphics[width=200pt]{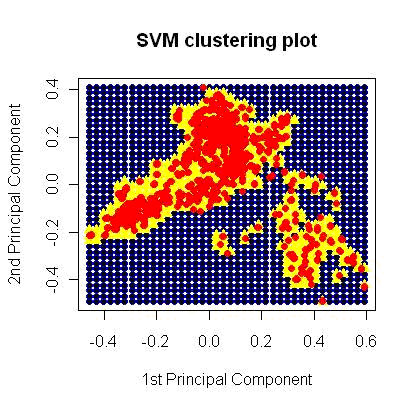}  }
\includegraphics[width=200pt]{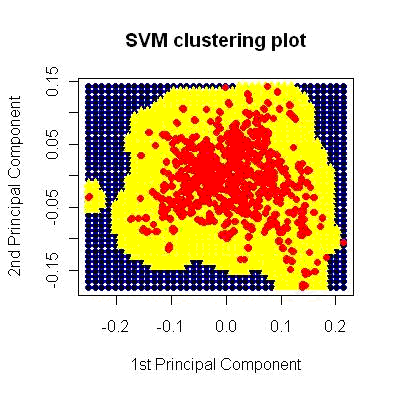} 
\caption{SVC using string kernel-constant (left), or using string kernel-spectrum (right).}
\label{figure10}
\end{figure}

If we pick two termes from our biology term data set : SK-constant ("inner coat", "in the mother cell") = 22, and SK-spectrum ("inner coat", "in the mother cell") = 15 ; another pair give SK-constant ("inner coat", "initiation of sporulation") = 27, and SK-spectrum ("inner coat", "initiation of sporulation") = 24. Variation between both pairs are not far according string kernels, though terms of the first pair are from one class (class 2) and the other pair compares terms from different classes (class 1 and class 2). We built a kernel matrix using both string kernels and achieved cluster labelling with this similarity information. Result is shown in Figure~\ref{figure10} :

Even if SK-constant shows some capability to isolate clusters, a big cluster contains 1600 items, hence 85\% of information. Such kernel is though challenging, perhaps including more lexical knowledge.

\subsection[Clustering capacity]{Clusterability of a SVC model}

Section~\ref{subsec:fram} presents a general framework of a kernel method. It does not mean any assumption about clustering but moreover about classification. Nevertheless SVC is not a new technique in itself. SVC has been seen as one-cluster discovery since a ball in the dual space is targeted. Hence it was described in detail for a long time as a one class approach applied to novelty detection when information is deviating from a block of well known information. In R-project, kernlab package \cite{Karatzoglou:2004, Karatzoglou:Meyer:Hornik:2006:JSSOBK:v15i09} implements novelty detection task. When running one-class kernel to our dataset it returns a model of 394 support vectors, with $nu = 0.2$ and cross-validation 0.205. Our observation is that SVC performs well cluster extraction (or labeling) from a 2-dimensional map, depending on existence condition of clusters. It means that data ought to be separable in the 2-d map. Separability can be managed by composition of a metric with a radial-based function over the whole input matrix dimensions. A possible explanation for capability of SVC to identify clusters is related to the same problem as trying to flatten the skin of an orange onto a tabletop. In this case, projection is a procedure to transform locations and features from the three-dimensional surface onto the two-dimensional paper in a defined and consistent way. The result is some slight bulges and a lot of gaps. The transformation of map information from a sphere to a flat sheet can be accomplished in many ways but no single projection can accurately portray area, shape, scale, and direction. SVC clustering takes origin from capacity within projections to distort.

\section[Conclusion]{Conclusion}

We developed, improved and applied a density and kernel based method called support vector clustering (SVC) we implemented as an R-project package (svcR). The package is available from the CRAN R project server (http://cran.r-project.org/ see Software, Packages; svcR version 1.4), and downloadable from the R graphical user interface (required R libraries : quadprog, ade4 and spdep). First we proved that mapping points in the data space to a grid and using the sphere radius from the attribute space and a k-nearest-neighbor approach improves time consumption for cluster labeling. In this sense, SVC can be seen as an efficient cluster extraction if clusters are separable in a 2-D map. Secondly we found a representation for term clustering using a mixed Jaccard-Radial base kernel and we proved its efficiency with SVC for term clustering in a natural language processing task as lexical classification (i.e. oriented ontology knowledge acquisition). Some investigations remain under R implementation to integrate C functions for matrix acquisition so as to make the toolkit more scalable in data size. Semantic and lexical-based kernels are promising for application in text mining frameworks. Yet it must understand how to select and integrate attributes for the description of terms. We aim at investigating in future work extraction of clusters over more than 2 dimensions, and test of robustness for non-separable data.

\section*{Acknowledgments}
Special thanks to Roy Varshavsky, Marine Campedel, Dunyon Lee and Olivier Chapelle for their discussion. The methodology discussed in this paper has been supported by the INRA-1077-SE grant from the French Institute for Agricultural Research (agriculture, food \& nutrition, environment and basic biology).

\nocite{*}
\bibliographystyle{plain}
\bibliography{Turenne}

\end{document}